\documentclass[runningheads]{llncs}
\usepackage{graphicx}
\usepackage{comment}
\usepackage{amsmath,amssymb} 
\usepackage{color}
\usepackage{multirow}
\usepackage[table, svgnames, dvipsnames]{xcolor}
\usepackage{enumitem}
\usepackage{subfigure}
\usepackage{caption}

\definecolor{citecolor}{RGB}{119,185,0}

\usepackage[breaklinks=true,letterpaper=true,colorlinks,citecolor=citecolor,bookmarks=false]{hyperref}

\newlength\savewidth\newcommand\shline{\noalign{\global\savewidth\arrayrulewidth
  \global\arrayrulewidth 1pt}\hline\noalign{\global\arrayrulewidth\savewidth}}

\begin{document}
\pagestyle{headings}
\mainmatter
\def\ECCVSubNumber{2974}  

\newcommand{\XY}[1]{{\color{cyan}[XY: #1]}}
\newcommand{\ZY}[1]{{\color{red}[ZY: #1]}}
\newcommand{\JK}[1]{{\color{magenta}[JK: #1]}}
\newcommand{\Yang}[1]{{\color{blue}[Yang: #1]}}

\renewcommand{\tabcolsep}{2.5pt}

\title{Joint Disentangling and Adaptation for Cross-Domain Person Re-Identification} 

\titlerunning{Joint Disentangling and Adaptation for Person ReID}
%
\author{Yang Zou\inst{1}\thanks{Work done during an internship at NVIDIA Research.} \quad 
Xiaodong Yang\inst{2} \quad 
Zhiding Yu\inst{2} \quad \\ 
B.V.K. Vijaya Kumar\inst{1} \quad Jan Kautz\inst{2}}
%
\authorrunning{Y. Zou, X. Yang, Z. Yu, B.V.K.V. Kumar, J. Kautz}
%
\institute{$^1$Carnegie Mellon University \quad $^2$NVIDIA}
\maketitle

\begin{abstract}
Although a significant progress has been witnessed in supervised person re-identification (re-id), it remains challenging to generalize re-id models to new domains due to the huge domain gaps. Recently, there has been a growing interest in using unsupervised domain adaptation to address this scalability issue. Existing methods typically conduct adaptation on the representation space that contains both id-related and id-unrelated factors, thus inevitably undermining the adaptation efficacy of id-related features. In this paper, we seek to improve adaptation by purifying the representation space to be adapted. To this end, we propose a joint learning framework that disentangles id-related/unrelated features and enforces adaptation to work on the id-related feature space exclusively. Our model involves a disentangling module that encodes cross-domain images into a shared appearance space and two separate structure spaces, and an adaptation module that performs adversarial alignment and self-training on the shared appearance space. The two modules are co-designed to be mutually beneficial. Extensive experiments demonstrate that the proposed joint learning framework outperforms the state-of-the-art methods by clear margins. 
\end{abstract}

\section{Introduction}
Person re-identification (re-id) is a task of retrieving the images that contain the person of interest across non-overlapping cameras given a query image.
It has been receiving lots of attention as a popular benchmark for metric-learning and found wide real applications such as smart cities~\cite{aicity20,pamtri,Tang_2019_CVPR,vehiclex}. Current state-of-the-art re-id methods predominantly hinge on deep convolutional neural networks (CNNs) and have considerably boosted re-id performance in the supervised learning scenario~\cite{sun2018beyond,attentive-siamese,zheng2019joint}. However, this idealistic closed-world setting postulates that training and testing data has to be drawn from the same camera network or the same domain, which rarely holds in real-world deployments. As a result, these re-id models usually encounter a dramatic performance degradation when deployed to new domains, mainly due to the great domain gaps between training and testing data, such as the changes of season, background, viewpoint, illumination, camera, etc. This largely restricts the applicability of such domain-specific re-id models, in particular, relabeling a large identity corpus for every new domain is prohibitively costly.

\begin{figure}[t]
	\centering
	\includegraphics[width=\linewidth]{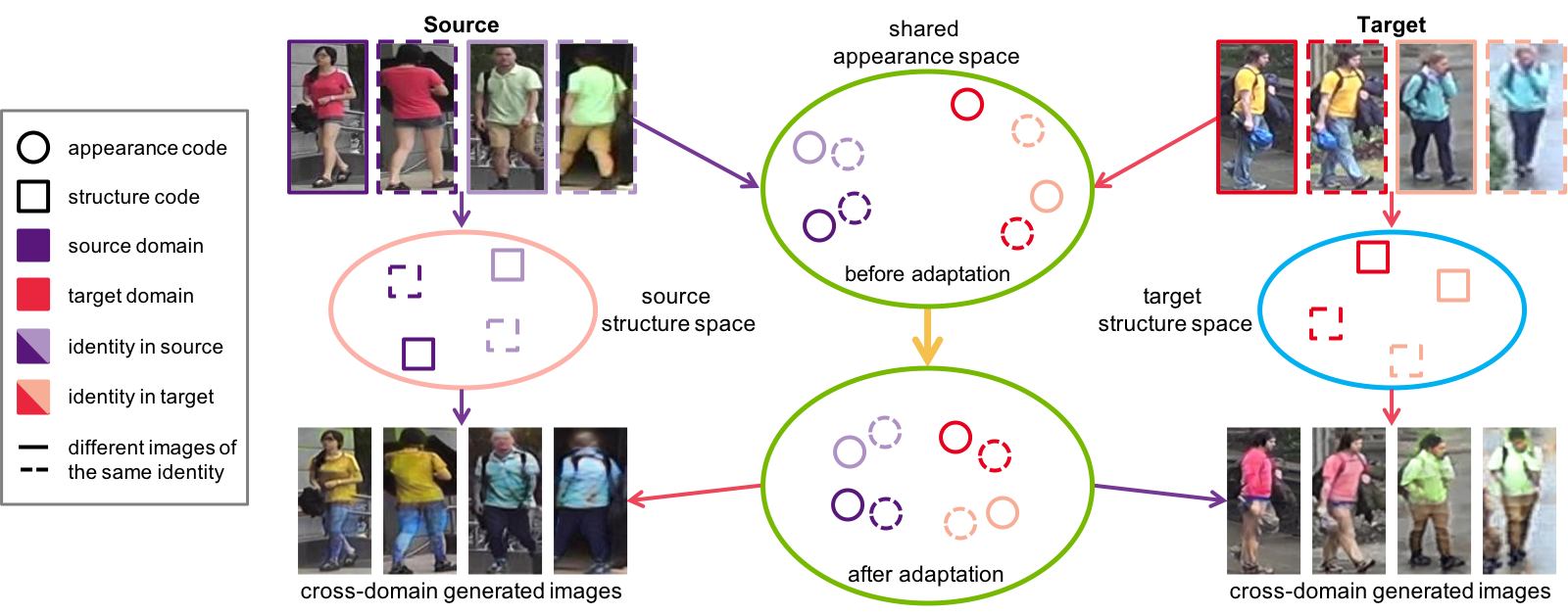}
	\caption{An overview of the proposed joint disentangling and adaptation framework. The disentangling module encodes images of two domains into a shared appearance space (id-related) and a separate source/target structure space (id-unrelated) via cross-domain image generation. Our adaptation module is exclusively conducted on the id-related feature space, encouraging the intra-class similarity and inter-class difference of the disentangled appearance features.} 
	\vspace{-10pt}
	\label{fig:teaser}
\end{figure}

To solve this problem, recent years have seen growing interests in person re-id under cross-domain settings. One popular solution to reduce the domain gap is unsupervised domain adaptation (UDA), which utilizes both labeled data in the source domain and unlabeled data in the target domain to improve the model performance in the target domain~\cite{hoffman2018cycada,Zou_2019_ICCV}. A fundamental design principle is to align feature distributions across domains to reduce the gap between source and target. A well-performing source model is expected to achieve similar performance in the target domain if the cross-domain gap is closed.



Compared to the conventional problems of UDA, such as image classification and semantic segmentation, person re-id is a more challenging open-set problem as two different domains contain disjoint or completely different identity class spaces. Recent methods mostly bridge the domain gap through adaptation at input-level and or feature-level. For input-level, the generative adversarial networks (GANs) are often utilized to transfer the holistic or factor-wise image style from source to target~\cite{deng2018image,adaptive-transfer}. Adaptation at feature-level often employs self-training or distribution distance minimization to enforce similar cross-domain distributions~\cite{linmulti,wang2018transferable}. Zhong et al.~\cite{zhunzhong2018eccv} combine the complementary benefits of both input-level and feature-level to further improve adaptation capability. 

However, a common issue behind these methods is that such adaptations typically operate on the feature space, which encodes both id-related and id-unrelated factors. Therefore, the adaptation of id-related features is inevitably interfered with and impaired by id-unrelated features, restricting the performance gain from UDA. 
Since cross-domain person re-id is coupled with both disentangling and adaptation problems, and existing methods mostly treat the two problems separately, it is important to come up with a principled framework that solves both issues together. Although disentangling has been studied for supervised person re-id in~\cite{eom2019learning,zheng2019joint}, it remains an open question how to integrate with adaptation, and it is under-presented in unsupervised cross-domain re-id as a result of the large domain gap and lack of target supervision.

In light of the above observation, we propose a joint learning framework that disentangles id-related/unrelated factors so that adaptation can be more effectively performed on the id-related space to prevent id-unrelated interference. Our work is partly inspired by DG-Net~\cite{zheng2019joint}, a recent supervised person re-id approach that performs within-domain image disentangling and leverages such disentanglement to augment training data towards better model training. 
We argue that successful cross-domain disentangling can create a desirable foundation for more targeted and effective domain adaptation.
We thus propose a cross-domain and cycle-consistent image generation with three latent spaces modeled by corresponding encoders to decompose source and target images. The latent spaces incorporate a \textbf{shared appearance} space that captures id-related features (i.e., appearance and other semantics), a \textbf{source structure} space and a \textbf{target structure} space that contain id-unrelated features (i.e., pose, position, viewpoint, background and other variations). We refer to the encoded features in the three spaces as codes. Our adaptation module 
is exclusively conducted in the shared appearance space, as illustrated in Figure~\ref{fig:teaser}. 

This design forms a joint framework that creates mutually beneficial cooperation between the disentangling and adaptation modules: (1) disentanglement leads to better adaptation as we can make the latter focus on id-related features and mitigate the interference of id-unrelated features, and (2) adaptation in turn improves disentangling as the shared appearance encoder gets enhanced during adaptation. We refer the proposed cross-domain joint disentangling and adaptation learning framework as \textbf{DG-Net++}. 

Our main contributions of this paper are summarized as follows. First, we propose a joint learning framework for unsupervised cross-domain person re-id to disentangle id-related/unrelated factors so that adaptation can be more effectively performed on the id-related space. Second, we introduce a cross-domain cycle-consistency paradigm 
to realize the desired disentanglement. Third, our disentangling and adaptation are co-designed to let the two modules mutually promote each other. Fourth, our approach achieves superior results on six benchmark pairs, largely pushing person re-id systems toward real-world deployment. Our code and model are available at~\url{https://github.com/NVlabs/DG-Net-PP}.

\begin{figure*}[!t]
	\centering
	\includegraphics[width=\textwidth]{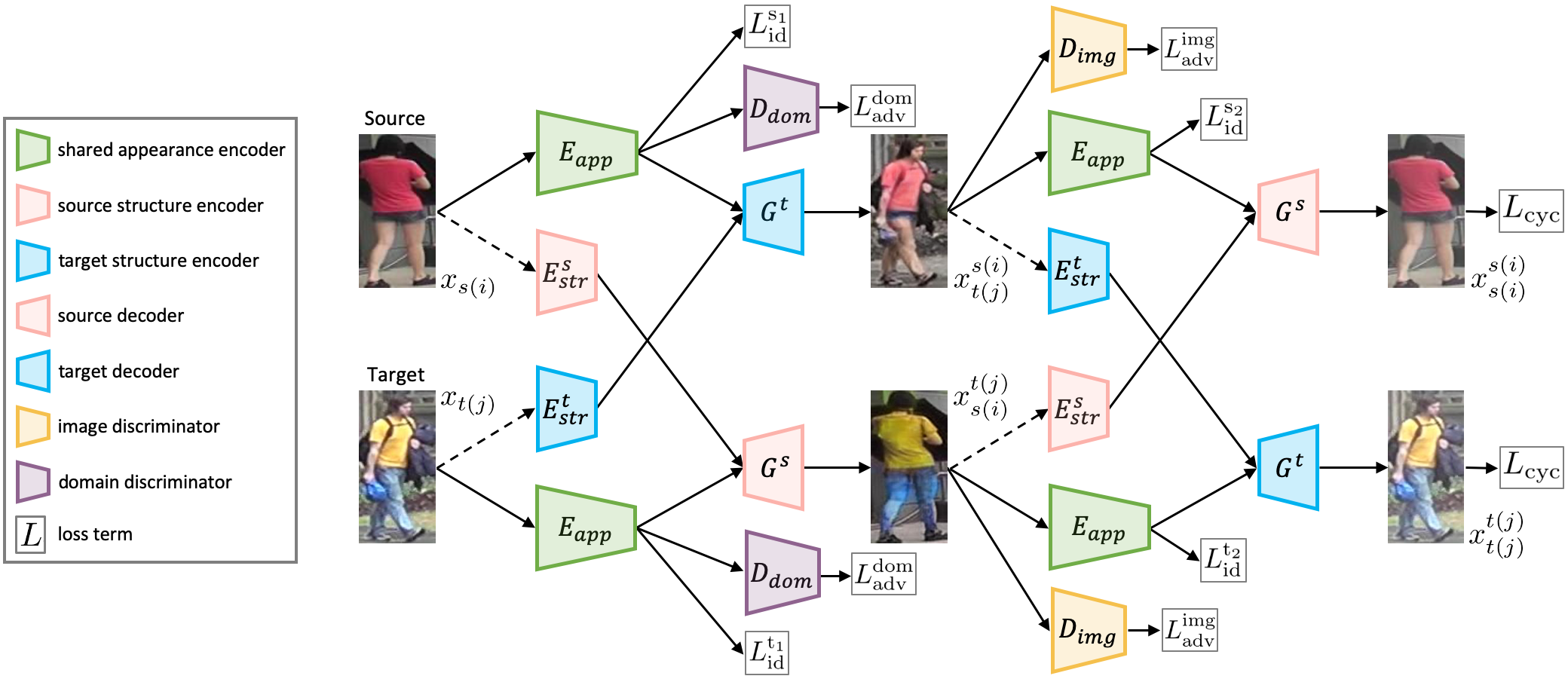}
	\caption{A schematic overview of the cross-domain cycle-consistency image generation. Our disentangling and adaptation modules are connected by the shared appearance encoder. The two domains also share the image and domain discriminators, but have their own structure encoders and decoders. A dashed line indicates that the input image to the source/target structure encoder is converted to gray-scale.}
	\vspace{-10pt}
	\label{fig:cycle}
\end{figure*}

\section{Related Work}

\textbf{Disentangling.} This task explores explanatory and independent factors among features in a representation. A generic framework combining deep convolutional auto-encoder with adversarial training is proposed in~\cite{mathieu2016disentangling} to disentangle hidden factors within a set of labeled observations. InfoGAN~\cite{chen2016infogan} and $\beta$-VAE~\cite{higgins2017beta} are introduced to learn interpretable factorized features in an unsupervised manner. A two-step disentanglement method~\cite{hadad2018two} is used to extract label relevant information for image classification. In~\cite{Huang_2018_ECCV,lee2018diverse}, images are decomposed to content and style information to serve image-to-image translation.


\noindent\textbf{Unsupervised domain adaptation.} UDA has been gaining increasing attention in image classification, object detection, and semantic segmentation. Based on the typical closed-set assumption that label classes are shared across domains, UDA methods can be roughly categorized as input-level and or feature-level adaptation. At input-level, models are usually adapted by training with style translated images~\cite{dundar2020domain,Huang_2018_ECCV,lee2018diverse}. Adaptation at feature-level often minimizes certain distance or divergence between source and target feature distributions, such as correlation~\cite{sun2016deep}, maximum mean discrepancy (MMD)~\cite{long2015learning}, sliced Wasserstein discrepancy~\cite{lee2019sliced}, and lifelong learning~\cite{chen2020automated}. Moreover, domain adversarial~\cite{hong2018conditional,tzeng2017adversarial} and self-training~\cite{chen2020angular,Ge2020Mutual,zou2018unsupervised,Zou_2019_ICCV} have also shown to be powerful feature-level alignment methods. CyCADA~\cite{hoffman2018cycada} adapts at both input-level and feature-level with the purpose of incorporating the effects of both.

\noindent\textbf{Person re-id.} A large family of person re-id focuses on supervised learning. They usually approach re-id as deep metric learning problems~\cite{Fan_2020_CVPR,hermans2017defense}, exploit pedestrian attributes as extra supervisions via multitask learning~\cite{su2016deep,wang2018transferable}, utilize part-based matching or ensembling to reduce intra-class variations~\cite{su2017pose,wei2017glad}, make use of human pose and parsing to facilitate local feature learning~\cite{semantic-parsing,zheng2020person}, or resort to generative models to augment training data~\cite{ge2018fd,zheng2019joint}. 
Although these methods have achieved tremendous progress in supervised setting, their performances degrade significantly on new domains.


Similar to the traditional problems of UDA, feature-level adaptation is widely used to seek source-target distribution alignment. In~\cite{li2019cross,linmulti}, feature adaptation is enforced by minimizing MMD between feature distributions in two domains. The self-training based methods also present promising results in~\cite{song2018unsupervised}. Another line is at input-level using GANs to transfer source images into target styles. An adaptive transfer method is developed in~\cite{adaptive-transfer} to decompose a holistic style to a set of imaging factors. Li et al.~\cite{li2019cross} propose to learn domain-invariant representation through pose-guided image translation. Chen et al.~\cite{yanbei-chen} present an instance-guided context rendering to enable supervised learning in target domain by transferring source person identities into target contexts. 

Although DG-Net++ inherits (and extends) the appearance and structure spaces of DG-Net~\cite{zheng2019joint}, there exist significant new designs in DG-Net++ to allow it to work for a very different problem. (1) DG-Net++ aims to address unsupervised cross-domain re-id, while DG-Net is developed under the fully supervised setting. (2) DG-Net++ is built upon a new cross-domain cycle-consistency scheme to disentangle id-related/unrelated factors without any target supervision. In comparison, DG-Net employs a within-domain disentanglement through latent code reconstruction with access to the ground truth identity. (3) DG-Net++ seamlessly integrates disentangling with adaptation in a unified manner to enable the two modules to mutually benefit each other, which is not considered in DG-Net. (4) DG-Net++ substantially outperforms DG-Net for unsupervised cross-domain re-id on six benchmark pairs. 

\section{Method}

As illustrated in Figure~\ref{fig:cycle}, DG-Net++ combines the disentangling and adaptation modules via the shared appearance encoder. We propose the cross-domain cycle-consistency generation to facilitate disentangling id-related (appearance) and id-unrelated (structure) factors. Our adaptation module involves adversarial alignment and self-training, which are co-designed with the disentangling module to target at id-related features and adapt more effectively.  

\subsection{Disentangling Module}

\textbf{Formulation.} 
We denote real images and labels in source domain as $X_{s}=\{x_{s(i)}\}_{i=1}^{N_s}$ and $Y_{s} = \{y_{s(i)}\}_{i=1}^{N_{s}}$, where $s$ indicates source domain, $N_{s}$ is the number of source images, $y_{s(i)} \in [1,K_{s}]$ and $K_{s}$ is the number of source identities. Similarly, $X_{t}=\{x_{t(i)}\}_{i=1}^{N_{t}}$ denotes $N_t$ real images in target domain $t$. Given a source image $x_{s(i)}$ and a target image $x_{t(j)}$, a new cross-domain synthesized image can be generated by swapping the appearance or structure codes between the two images. As shown in Figure~\ref{fig:cycle}, the disentangling module consists of a shared appearance encoder $E_{app}: x \rightarrow \nu$, a source structure encoder $E_{str}^s: x_{s(i)} \rightarrow \tau_{s(i)}$, a target structure encoder $E_{str}^t: x_{t(j)} \rightarrow \tau_{t(j)}$, a source decoder $G^s: (\nu_{t(j)}, \tau_{s(i)}) \rightarrow x_{s(i)}^{t(j)}$, a target decoder $G^t: (\nu_{s(i)}, \tau_{t(j)}) \rightarrow x_{t(j)}^{s(i)}$, an image discriminator $D_{img}$ to distinguish between real and synthesized images, and a domain discriminator $D_{dom}$ to distinguish between source and target domains. Note: for synthesized images, we use superscript to indicate the real image providing appearance code and subscript to denote the one giving structure code; for real images, they only have subscript as domain and image index. 
Our adaptation and re-id are conducted using the appearance codes. 

\noindent\textbf{Cross-domain generation.} We introduce cross-domain cycle-consistency image generation to enforce disentangling between appearance and structure factors. Given a pair of source and target images, we first swap their appearance or structure codes to  synthesize new images. Since there exists no ground-truth supervision for the synthetic images, we take advantage of cycle-consistency self-supervision to reconstruct the two real images by swapping the appearance or structure codes extracted from the synthetic images. As demonstrated in Figure~\ref{fig:cycle}, given a source image $x_{s(i)}$ and a target image $x_{t(j)}$, the synthesized images $x_{t(j)}^{s(i)} = G^t(\nu_{s(i)}, \tau_{t(j)})$ and $x_{s(i)}^{t(j)} = G^s(\nu_{t(j)}, \tau_{s(i)})$ are required to respectively preserve the corresponding appearance and structure codes from $x_{s(i)}$ and $x_{t(j)}$ to be able to reconstruct the two original real images: 
\begin{equation}
\begin{aligned}
    L_{\mathrm{cyc}} = \hspace{1mm} &\mathbb{E}\left[ \left\lVert x_{s(i)} - G^{s}(E_{app}(x^{s(i)}_{t(j)}), E^s_{str}(x^{t(j)}_{s(i)})) \right\rVert_1 \right] + \\
    & \mathbb{E}\left[ \left\lVert x_{t(j)} - G^{t}(E_{app}(x_{s(i)}^{t(j)}), E_{str}^{t}(x_{t(j)}^{s(i)})) \right\rVert_1 \right].
\end{aligned}
\end{equation}
\noindent With the identity labels available in source domain, we then explicitly enforce the shared appearance encoder to capture the id-related information by using the identification loss:
\begin{equation}
    L^{\mathrm{s_1}}_{\mathrm{id}} = \mathbb{E}[-\log(p(y_{s(i)}|x_{s(i)}))].
\end{equation}
\noindent where $p(y_{s(i)}|x_{s(i)})$ is the predicted probability that $x_{s(i)}$ belongs to the ground-truth label $y_{s(i)}$. We also apply the identification loss on the synthetic image that retains the appearance code from source image to keep identity consistency: 
\begin{align}
    L^{\mathrm{s_2}}_{\mathrm{id}} = \mathbb{E}[-\log(p(y_{s(i)}|x_{t(j)}^{s(i)}))].
\end{align}
\noindent where $p(y_{s(i)}|x_{t(j)}^{s(i)})$ is the predicted probability of $x_{t(j)}^{s(i)}$ belonging to the ground-truth label $y_{s(i)}$ of $x_{s(i)}$.
In addition, we employ adversarial loss to match the distributions between the synthesized images and the real data:
\begin{equation}
\begin{aligned}
L^{\mathrm{img}}_{\mathrm{adv}} = \hspace{1mm} 
& \mathbb{E}\left[\log D_{img}(x_{s(i)}) + \log(1 - D_{img}(x^{s(i)}_{t(j)})\right] +\\
& \mathbb{E}\left[\log D_{img}(x_{t(j)}) + \log(1 - D_{img}(x^{t(j)}_{s(i)})\right].
\end{aligned}
\end{equation}
\noindent Note that the image discriminator $D_{img}$ is shared across domains to force the synthesized images to be realistic regardless of domains. This can indirectly drive the shared appearance encoder to learn domain-invariant features. Apart from the cross-domain generation, our disentangling module is also flexible to incorporate the within-domain generation as~\cite{zheng2019joint}, which can be used to further stabilize and regulate the within-domain disentanglement. 


\subsection{Adaptation Module}

\textbf{Adversarial alignment.} Although the weights of appearance encoder are shared between source and target domains, the appearance representations across domains are still not ensured to have similar distributions. To encourage the alignment of appearance features in two domains, we introduce a domain discriminator $D_{dom}$, which aims to distinguish the domain membership of the encoded appearance codes $\nu_{s(i)}$ and $\nu_{t(j)}$. During adversarial training, the shared appearance encoder learns to produce appearance features of which domain membership cannot be differentiated by $D_{dom}$, such that the distance between cross-domain appearance feature distributions can be reduced. We express this domain appearance adversarial alignment loss as:
\begin{equation}
\begin{aligned}
L^{\mathrm{dom}}_{\mathrm{adv}} = \hspace{1mm} 
& \mathbb{E}\left[\log D_{dom}(\nu_{s(i)}) + \log(1 - D_{dom}(\nu_{t(j)})\right] + \\
& \mathbb{E}\left[\log D_{dom}(\nu_{t(j)}) + \log(1 - D_{dom}(\nu_{s(i)})\right].
\end{aligned}
\end{equation}

\noindent\textbf{Self-training.}
In addition to the global feature alignment imposed by the above domain adversarial loss, we incorporate self-training in the adaptation module. Essentially, self-training with identification loss is an entropy minimization process that gradually reduces intra-class variations. It implicitly closes the cross-domain feature distribution distance in the shared appearance space, and meanwhile encourages discriminative appearance feature learning.

We iteratively generate a set of pseudo-labels $\hat{Y}_{t} = \{\hat{y}_{t(j)}\}$ based on the reliable identity predictions in target domain, and refine the network using the pseudo-labeled target images. Note the numbers of pseudo-identities and labeled target images may change during self-training. In practice, the pseudo-labels are produced by clustering the target features that are extracted by the shared appearance encoder $E_{app}$. We assign the same pseudo-label to the samples within the same cluster. We adopt an affinity based clustering method DBSCAN~\cite{ester1996density} that has shown promising results in re-id. We utilize the K-reciprocal encoding~\cite{zhong2017re} to compute pairwise distances, and update pseudo-labels every two epochs. With the pseudo-labels obtained by self-training in target domain, we apply the identification loss on the shared appearance encoder:
\begin{equation}
L^{\mathrm{t_1}}_{\mathrm{id}} = \mathbb{E}[-\log(p(\hat{y}_{t(j)}|x_{t(j)}))].
\end{equation}
\noindent where $p(\hat{y}_{t(j)}|x_{t(j)})$ is the predicted probability that $x_{t(j)}$ belongs to the pseudo-label $\hat{y}_{t(j)}$. We furthermore enforce the identification loss with pseudo-label on the synthetic image that reserves the appearance code from target image to keep pseudo-identity consistency:
\begin{align}
    L^{\mathrm{t_2}}_{\mathrm{id}} = \mathbb{E}[-\log(p(\hat{y}_{t(j)}|x^{t(j)}_{s(i)}))].
\end{align}
\noindent where $p(\hat{y}_{t(j)}|x^{t(j)}_{s(i)})$ is the predicted probability of $x^{t(j)}_{s(i)}$ belonging to the pseudo-label $\hat{y}_{t(j)}$ of $x_{t(j)}$. Overall, adaptation with self-training encourages the shared appearance encoder to learn both domain-invariant and discriminative features that can generalize and facilitate re-id in target domain.

\subsection{Discussion}
Our disentangling and adaptation are co-designed to let the two modules positively interact with each other. On the one hand, \textbf{disentangling promotes adaptation}. Based on the cross-domain cycle-consistency image generation, our disentangling module learns detached appearance and structure factors with explicit and explainable meanings, paving the way for adaptation to exclude id-unrelated noises and specifically operate on id-related features. With the help of sharing appearance encoder, the discrepancy between cross-domain feature distributions can be reduced. Also the adversarial loss for generating realistic images across domains encourages feature alignment through the shared image discriminator. On the other hand, \textbf{adaptation facilitates disentangling}. In addition to globally close the distribution gap, the adversarial alignment by the shared domain discriminator helps to find the common appearance embedding that can assist disentangling appearance and structure features. Besides implicitly aligning cross-domain features, the self-training with the identification loss supports disentangling since it forces the appearance features of different identities to stay apart while reduces the intra-class variation of the same identity. 
Therefore, through the adversarial loss and identification loss via self-training, the appearance encoder is enhanced in the adaptation process, and a better appearance encoder generates better synthetic images, eventually leading to the improvement of the disentangling module. 



\subsection{Optimization}

We jointly train the shared appearance encoder, image discriminator, domain discriminator, as well as source and target structure encoders, and source and target decoders to optimize the total objective, which is a weighted sum of the following loss terms:
\begin{equation}
\begin{aligned}
&L_{\mathrm{total}}(E_{app}, D_{img}, D_{dom}, E_{str}^{s}, E_{str}^{t}, G^s, G^t) = \\
& \lambda_{\mathrm{cyc}} L_{\mathrm{cyc}} + L^{\mathrm{s_1}}_{\mathrm{id}} + L^{\mathrm{t_1}}_{\mathrm{id}} +\lambda_{\mathrm{id}} L^{\mathrm{s_2}}_{\mathrm{id}} +
\lambda_{\mathrm{id}} L^{\mathrm{t_2}}_{\mathrm{id}} +
L^{\mathrm{img}}_{\mathrm{adv}} + L^{\mathrm{dom}}_{\mathrm{adv}}.
\end{aligned}
\label{eq:total}
\end{equation}
\noindent where $\lambda_{\mathrm{cyc}}$ and $\lambda_{\mathrm{id}}$ are the weights to control the importance of cross-domain cycle-consistent self-supervision loss and identification loss on synthesized images. Following the common practice in image-to-image translations~\cite{Huang_2018_ECCV,lee2018diverse,zhu2017unpaired}, we set a large weight $\lambda_{\mathrm{cyc}} = 2$ for $L_{\mathrm{cyc}}$. As the quality of cross-domain synthesized images is not great at the early stage of training, the two losses $L^{\mathrm{s_2}}_{\mathrm{id}}$ and $L^{\mathrm{t_2}}_{\mathrm{id}}$ on such images would make training unstable, so we use a relatively small weight $\lambda_{\mathrm{id}} = 0.5$. We fix the weights during the entire training process in all experiments. We first warm up $E_{app}$, $E_{str}^s$, $G^s$ and $D_{img}$ with the disentangling module in source domain for $100$K iterations, then bring in the adversarial alignment to train the whole network for another $50$K before self-training. In the process of self-training, all components are co-trained, and the pseudo-labels are updated every two epochs. We follow the alternative updating policy in training GANs to alternatively train $E_{app}$, $E_{str}^s$, $E_{str}^t$, $G^s$, $G^t$, and $D_{img}$, $D_{dom}$.


\section{Experiments}
We evaluate the proposed framework DG-Net++ following the standard experimental protocols on six domain pairs formed by three benchmark datasets: Market-1501~\cite{zheng2015scalable}, DukeMTMC-reID~\cite{dukemtmc} and MSMT17~\cite{wei2018person}. We report comparisons to the state-of-the-art methods and provide in-depth analysis. A variety of ablation studies are performed to understand the contributions of each individual component in our approach. The qualitative results of cross-domain image generation are also presented. Extensive evaluations reveal that our approach consistently produces realistic cross-domain images, and more importantly, outperforms the competing algorithms by clear margins over all benchmarks.   

\subsection{Implementation Details}

We implement our framework in PyTorch. In the following descriptions, we use channel$\times$height$\times$width to denote the size of feature maps. 
(\textbf{1}) $E_{app}$ is modified from ResNet50~\cite{he2016deep} and  pre-trained on ImageNet~\cite{imagenet}. Its global average pooling layer and fully-connected layer are replaced with a max pooling layer that outputs the appearance code $\nu$ in $2048\times4\times1$, which is in the end mapped to a 1024-dim vector to perform re-id. (\textbf{2}) $E_{str}^s$ and $E_{str}^t$ share the same architecture with four convolutional layers followed by four residual blocks~\cite{he2016deep}, and output the source/target structure code $\tau$ in $128\times64\times32$. (\textbf{3}) $G^s$ and $G^t$ use the same decoding scheme to process the source/target code $\tau$ through four residual blocks and four convolutional layers. And each residual block includes two adaptive instance normalization layers~\cite{huang2017arbitrary} to absorb the appearance code $\nu$ as scale and bias parameters. (\textbf{4}) $D_{img}$ follows the popular multi-scale PatchGAN~\cite{isola2017image} at three different input scales: $64\times32$, $128\times64$, and $256\times128$. (\textbf{5}) $D_{dom}$ is a multi-layer perceptron containing four fully-connected layers to map the appearance code $\tau$ to a domain membership. 
(\textbf{6}) For training, input images are resized to $256\times128$. We use SGD to train $E_{app}$, and Adam~\cite{kingma2014adam} to optimize $E_{str}^s$, $E_{str}^t$, $G^s$, $G^t$, $D_{img}$, $D_{dom}$. (\textbf{7}) For generating pseudo-labels with DBSCAN in self-training, we set the neighbor maximum distance to $0.45$ and the minimum number of points required to form a dense region to $7$. (\textbf{8}) At test time, our re-id model only involves $E_{app}$, which has a comparable network capacity to most re-id models using ResNet50 as a backbone. We use the 1024-dim vector output by $E_{app}$ as the final image representation. 

\subsection{Quantitative Results}
\noindent\textbf{Comparison with the state-of-the-art.} We extensively evaluate DG-Net++ on six cross-domain pairs among three benchmark datasets with a variety of competing algorithms. Table~\ref{tab:du2ma} shows the comparative results on the six cross-domain pairs. In particular, compared to the second best methods, we achieve the state-of-the-art results with considerable margins of 10.4\%, 3.4\%, 8.9\%, 8.8\%, 24.5\%, 5.0\% mAP and 5.9\%, 2.1\%, 16.8\%, 16.6\%, 14.6\%, 3.2\% Rank$@1$ on Market $\rightarrow$ Duke, Duke $\rightarrow$ Market, Market $\rightarrow$ MSMT, Duke $\rightarrow$ MSMT, MSMT $\rightarrow$ Duke, MSMT $\rightarrow$ Market, respectively. Moreover, DG-Net++ is found to even outperform or approach some recent supervised re-id methods \cite{huang2018multi,liu2018pose,sun2017svdnet,unlabeld-samples,zheng2018pedestrian} that have access to the full labels of the target domain. 

These superior performances collectively and clearly show the advantages of the joint disentangling and adaptation design, which enables more effective adaptation in the disentangled id-related feature space and presents strong cross-domain adaptation capability. 
Additionally, we emphasize that the disentangling module in DG-Net++ is orthogonal and applicable to other adaptation methods without considering feature disentangling. Overall, our proposed cross-domain disentangling provides a better foundation to allow for more effective cross-domain re-id adaptation. Other adaptation methods, such as some recent approaches~\cite{Fu_2019_ICCV,Ge2020Mutual,zhong2019invariance}, can be readily applied to the disentangled id-related feature space, and their performances may even be boosted further.

\begin{table}[t]
\centering
\begin{tabular*}{\linewidth}{l|cccc|cccc}
\shline
\multirow{2}{*}{Methods} & \multicolumn{4}{c|}{Market-1501 $\rightarrow$ DukeMTMC-reID} & \multicolumn{4}{c}{DukeMTMC-reID $\rightarrow$ Market-1501} \\
  & Rank$@1$ & Rank$@5$ & Rank$@10$ & mAP & Rank$@1$ & Rank$@5$ & Rank$@10$ & mAP\\ \hline
SPGAN \cite{deng2018image} & 41.1 & 56.6 & 63.0 & 22.3  & 51.5 & 70.1 & 76.8 & 22.8 \\
AIDL \cite{wang2018transferable} & 44.3 & 59.6 & 65.0 & 23.0 & 58.2 & 74.8 &  81.1 & 26.5 \\
MMFA \cite{linmulti} & 45.3 & 59.8 & 66.3 & 24.7  & 56.7 & 75.0 & 81.8 & 27.4 \\
HHL \cite{zhunzhong2018eccv} & 46.9 & 61.0 & 66.7 & 27.2 & 62.2 & 78.8 & 84.0 & 31.4 \\
CAL \cite{Qi_2019_ICCV} & 55.4 & - & - & 36.7 & 64.3 & - & - & 34.5\\
ARN \cite{li2018adaptation} & 60.2 & 73.9 & 79.5 & 33.4 & 70.3 & 80.4 & 86.3 & 39.4\\
ECN \cite{zhong2019invariance} & 63.3 & 75.8 & 80.4 & 40.4 & 75.1 & 87.6 & 91.6 & 43.0 \\
PDA \cite{li2019cross} & 63.2 & 77.0 & 82.5 & 45.1 & 75.2 & 86.3 & 90.2 & 47.6 \\
CR-GAN \cite{yanbei-chen} & 68.9 & 80.2 & 84.7 & 48.6 & 77.7 & 89.7 & 92.7 & 54.0 \\
IPL \cite{song2018unsupervised} & 68.4 & 80.1 & 83.5 & 49.0 & 75.8 & 89.5 & 93.2 & 53.7 \\
SSG \cite{Fu_2019_ICCV} & 73.0 & 80.6 & 83.2 & 53.4 & 80.0 & 90.0 & 92.4 & 58.3
\\ \hline
DG-Net++ & \textbf{78.9} & \textbf{87.8} & \textbf{90.4} & \textbf{63.8} & \textbf{82.1} & \textbf{90.2} & \textbf{92.7} & \textbf{61.7}\\
\hline \hline
\multirow{2}{*}{Methods} & \multicolumn{4}{c|}{Market-1501 $\rightarrow$ MSMT17} & \multicolumn{4}{c@{}}{DukeMTMC-reID $\rightarrow$ MSMT17}  \\
 & Rank$@1$ & Rank$@5$ & Rank$@10$ & mAP & Rank$@1$ & Rank$@5$ & Rank$@10$ & mAP \\ \hline
 PTGAN \cite{wei2018person} & 10.2 & - & 24.4 & 2.9 & 11.8 & - & 27.4 & 3.3 \\
 ENC \cite{zhong2019invariance} & 25.3 & 36.3 & 42.1 & 8.5 & 30.2 & 41.5 & 46.8 & 10.2 \\
 SSG \cite{Fu_2019_ICCV} & 31.6 & - & 49.6 & 13.2 & 32.2 & - & 51.2 & 13.3 \\ \hline
DG-Net++ & \textbf{48.4} & \textbf{60.9} & \textbf{66.1} & \textbf{22.1} & \textbf{48.8} & \textbf{60.9} & \textbf{65.9} & \textbf{22.1} \\ \hline \hline
\multirow{2}{*}{Methods} & \multicolumn{4}{c|}{MSMT17 $\rightarrow$ Market-1501}  & \multicolumn{4}{c}{MSMT17 $\rightarrow$ DukeMTMC-reID}\\
 & Rank$@1$ & Rank$@5$ & Rank$@10$ & mAP & Rank$@1$ & Rank$@5$ & Rank$@10$ & mAP \\ \hline
 PAUL \cite{yang2019patch} & 68.5 & - & - & 40.1 & 72.0 & - & - & 53.2 \\ \hline
DG-Net++ & \textbf{83.1} & \textbf{91.5} & \textbf{94.3} & \textbf{64.6} & \textbf{75.2} & \textbf{73.6} & \textbf{86.9} & \textbf{58.2} \\
\shline
\end{tabular*}
\vspace{1mm}
\caption{Comparison with the state-of-the-art unsupervised cross-domain re-id methods on the six cross-domain benchmark pairs.}
\vspace{-8pt}
\label{tab:du2ma}
\end{table}

\noindent\textbf{Ablation study.} We perform a variety of ablation experiments primarily on the two cross-domain pairs: Market $\rightarrow$ Duke and Duke $\rightarrow$ Market to evaluate the contribution of each individual component in DG-Net++. As shown in Table~\ref{tab:ablation}, our baseline is an ImageNet pre-trained ResNet50 that is trained on the source domain and directly transferred to the target domain. By just using the proposed disentangling module, our approach can boost the baseline performance by 4.9\%, 11.8\% mAP and 7.1\%, 10.4\% Rank$@$1 respectively on the two cross-domain pairs. Note this improvement is achieved without using any adaptations. This suggests that by only removing the id-unrelated features through disentangling, the cross-domain discrepancy has already been reduced since the id-unrelated noises largely contribute to the domain gap. 
Based on the disentangled id-related features, either adversarial alignment or self-training consistently provides clear performance gains. By combining both, our full model obtains the best performances that are substantially improved over the baseline results. 

\begin{figure}
\centering     
\subfigure[]{\label{fig:ablation_D}\includegraphics[width=0.475\textwidth]{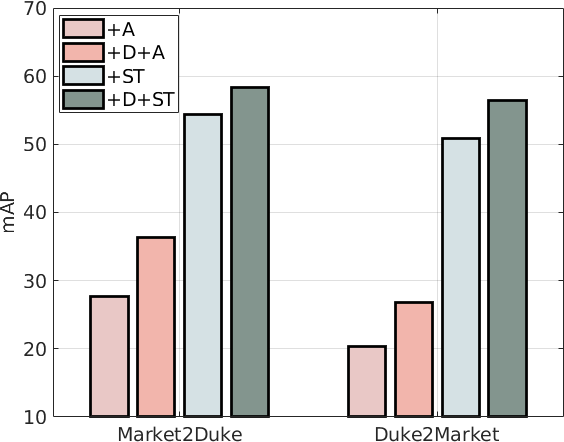}}
\subfigure[]{\label{fig:curve}\includegraphics[width=0.475\textwidth]{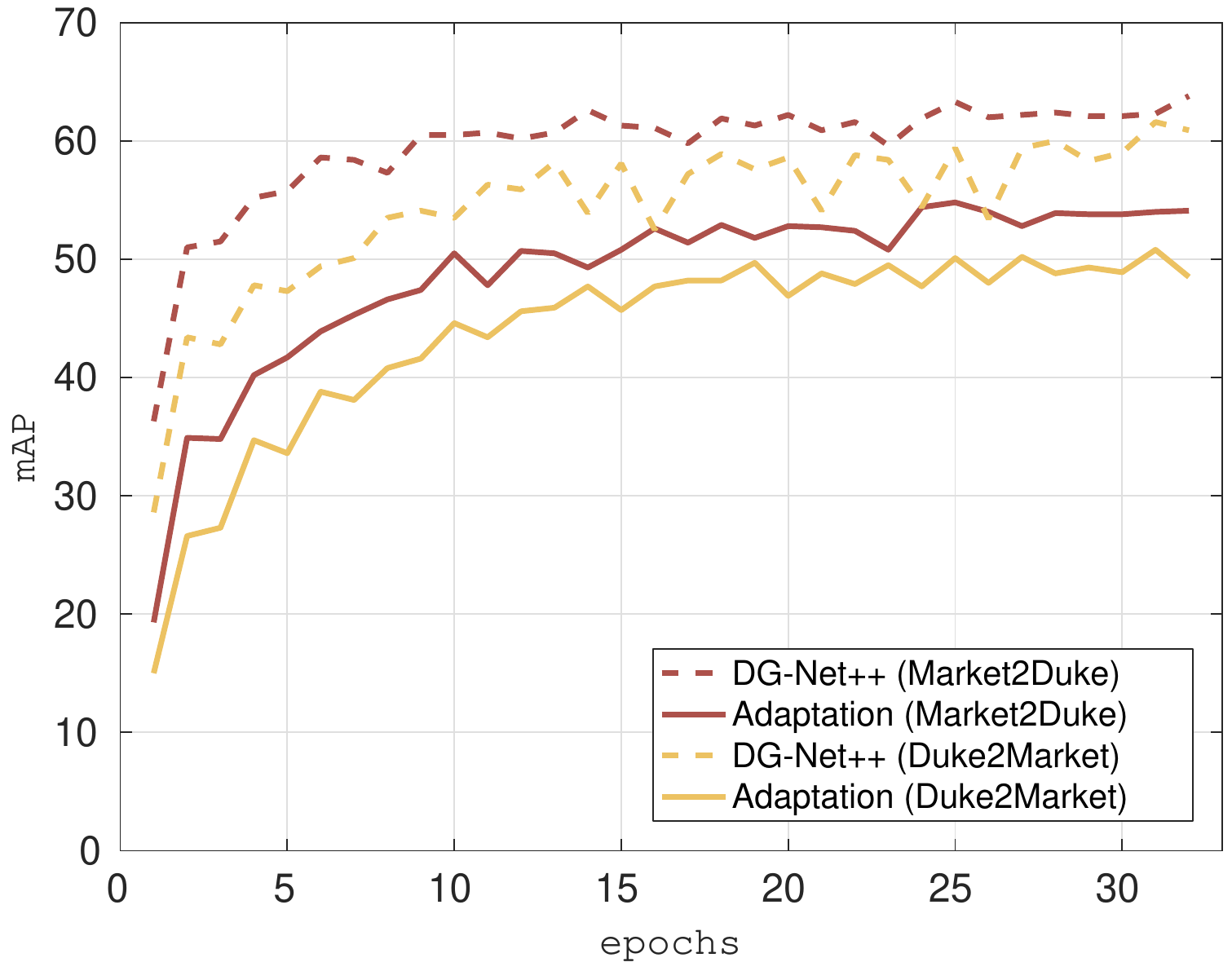}}
\vspace{-.35cm}
\caption{(a) Improvements of disentangling to adaptation in DG-Net++. ``A'': adversarial alignment, ``ST'': self-training, and ``D'': disentangling. (b) Comparison of the training processes between our full model and the adaptation (self-training) alone model on the two cross-domain pairs.}
\end{figure}

Next we study the gains of disentangling to adaptation in DG-Net++. As shown in Figure~\ref{fig:ablation_D}, compared with the space entangled with both id-related and id-unrelated factors, in the disentangled id-related space, adversarial alignment can be conducted more effectively with 8.6\% and 6.4\% mAP improvements on Market $\rightarrow$ Duke and Duke $\rightarrow$ Market, respectively. A similar observation can also be found for self-training. In comparison to self-training only, disentangling largely boosts the performance by 4.0\% and 5.7\% mAP on the two cross-domain pairs. This strongly indicates the advantages of disentangling to enable more effective adaptation in the separated id-related space.

\begin{table}[t]
\centering
\begin{tabular*}{\linewidth}{l|cccc|cccc}
\shline \
\multirow{2}{*}{Methods} & \multicolumn{4}{c|}{Market-1501 $\rightarrow$ DukeMTMC-reID} & \multicolumn{4}{c}{DukeMTMC-reID $\rightarrow$ Market-1501}\\
 & Rank$@1$ & Rank$@5$ & Rank$@10$ & mAP & Rank$@1$ & Rank$@5$ & Rank$@10$ & mAP \\ \hline
Baseline & 37.4 & 52.4 & 58.4 & 19.3 & 39.7 & 57.9 & 64.3 & 15.0 \\
+A+ST & 71.4 & 81.8 & 85.7 & 57.5 & 75.7 & 86.4 & 90.1 & 57.1 \\
+D & 44.5 & 60.6 & 66.7 & 24.2 & 50.1 & 68.0 & 73.9 & 26.8 \\
+D+A & 53.2 & 68.7 & 73.8 & 36.3 & 52.2 & 70.7 & 77.0 & 28.6 \\
+D+ST & 74.2 & 82.8 & 86.5 & 58.4 & 78.0 & 87.1 & 90.3 & 56.5 \\
+D+A+ST & \textbf{78.9} & \textbf{87.8} & \textbf{90.4} & \textbf{63.8} & \textbf{82.1} & \textbf{90.2} & \textbf{92.7} & \textbf{61.7}\\ 
\shline
\end{tabular*}
\vspace{1mm}
\caption{Ablation study on two cross-domain pairs: Market $\rightarrow$ Duke and Duke $\rightarrow$ Market. We use ``D'' to denote disentangling, ``A'' to adversarial alignment, and ``ST'' to self-training.}
\vspace{-10pt}
\label{tab:ablation}
\end{table}


To better understand the learning behavior of DG-Net++, we plot the training curves on the two cross-domain pairs in Figure~\ref{fig:curve}. Our full model consistently outperforms the self-training alone model by large margins during the training process thanks to the merits that the adaptation can be more effectively performed on the disentangled id-related space in our full model. In addition, as shown in the figure, the training curves are overall stable with slight fluctuations after 13 epochs, and we argue that such a stable learning behavior is quite desirable for model selection in the unsupervised cross-domain scenario where the target supervision is not available. 

\noindent\textbf{Comparison with DG-Net.} To validate the superiority of DG-Net++ over DG-Net for unsupervised cross-domain adaptation, we conduct further ablation study on Market $\rightarrow$ Duke. (1) Based on DG-Net trained in source domain, we perform self-training with the trained model, i.e., the appearance encoder. It achieves 54.6\% mAP, 9.2\% inferior to 63.8\% mAP of DG-Net++. This shows the necessity of joint disentangling and adaptation for cross-domain re-id. (2) We perform a semi-supervised training for DG-Net on two domains, where self-training is introduced to supervise the appearance encoder in target domain. It achieves 52.9\% mAP, 10.9\% inferior to DG-Net++. Note this result is even worse than self-training with only the appearance encoder (54.6\%). This suggests that an inappropriate design of disentangling (the within-domain disentangling of DG-Net) can harm adaptation. In summary, DG-Net is designed to work on a single domain, while the proposed disentangling of DG-Net++ is vital for a joint disentangling and adaptation in cross-domain.


\noindent\textbf{Sensitivity analysis.} We also study how sensitive the re-id performance is to the two important hyper-parameters in Eq.~\ref{eq:total}: one is $\lambda_{\mathrm{cyc}}$, the weight to control the importance of $L_{\mathrm{cyc}}$; the other is $\lambda_{\mathrm{id}}$ to weight the identification losses $L^{\mathrm{s_2}}_{\mathrm{id}}$ and $L^{\mathrm{t_2}}_{\mathrm{id}}$ on the synthesized images of source and target domains. This analysis is conducted on Market $\rightarrow$ Duke. Figure~\ref{fig:hyper} demonstrates that the re-id performances are overall stable and there are only slight variations when $\lambda_{\mathrm{cyc}}$ varies from 1 to 4 and $\lambda_{\mathrm{id}}$ from 0.25 to 1. Thus, our model is not sensitive to the two hyper-parameters, and we set $\lambda_{\mathrm{cyc}} = 2$ and $\lambda_{\mathrm{id}} = 0.5$ in all experiments.

\begin{figure}
\centering     
\subfigure[]{\label{fig:hyper}\includegraphics[width=0.44\textwidth]{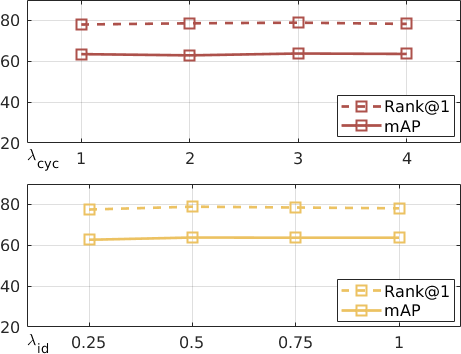}}~~
\subfigure[]{\label{fig:visual_ablation}\includegraphics[width=0.55\textwidth]{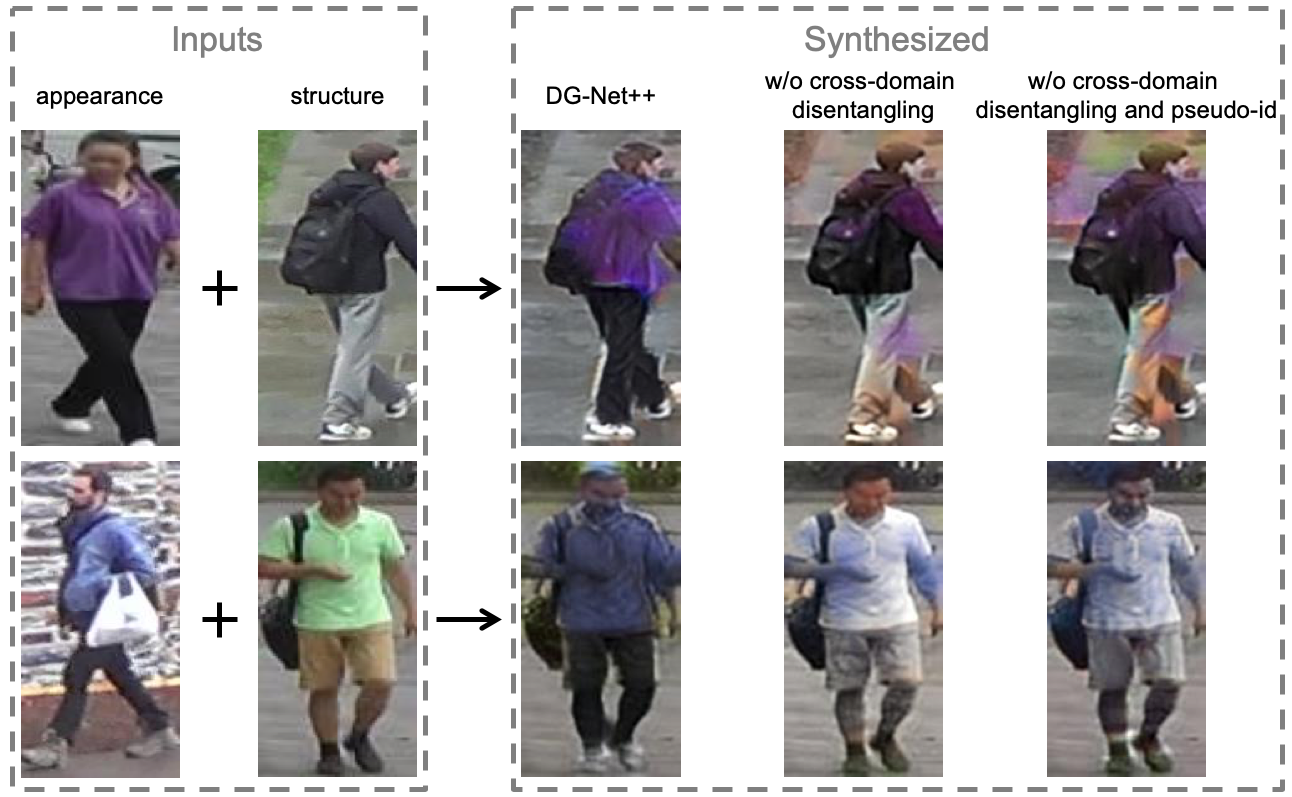}}
\vspace{-.35cm}
\caption{(a) Analysis of the influence of hyper-parameters $\lambda_{\mathrm{cyc}}$ and $\lambda_{\mathrm{id}}$ on Market $\rightarrow$ Duke. (b) Comparison of the synthesized images by our full model, removing cross-domain disentangling, and further removing pseudo-identity supervision. We use source appearance and target structure in the first row, and target appearance and source structure in the second row.}
\vspace{-10pt}
\end{figure}


\subsection{Qualitative Results}

\begin{figure*}[!t]
	\centering
    \includegraphics[width=\textwidth]{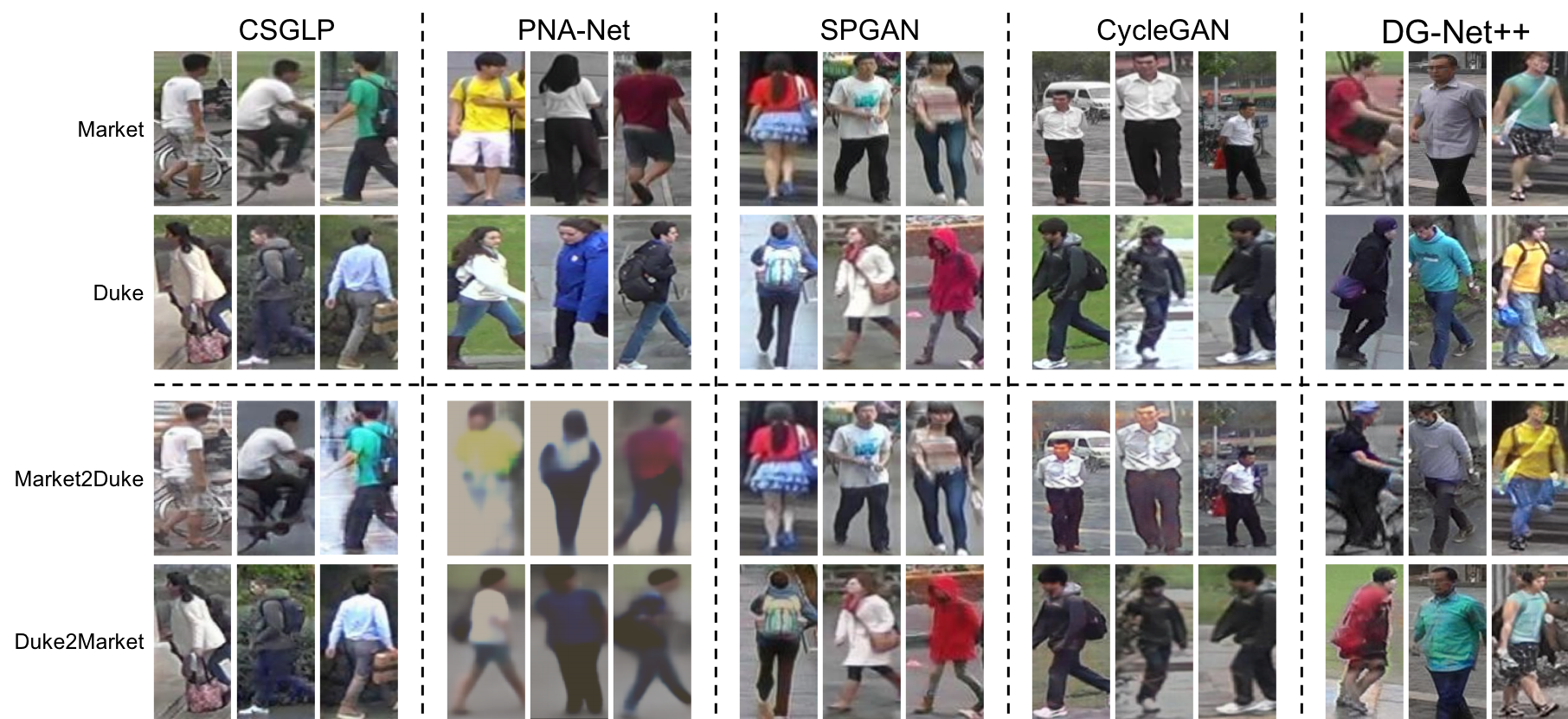}
    \caption{Comparison of the generated images across two cross-domains between Market and Duke of different methods including CycleGAN \cite{zhu2017unpaired}, SPGAN \cite{deng2018image}, PNA-Net \cite{li2019cross}, CSGLP \cite{ren2019domain}, and our approach DG-Net++.  Please attention to both foreground and background of the synthetic images.}
    \vspace{-10pt}
    \label{fig:translation}
\end{figure*}

\begin{figure*}[!h]
	\centering
	\includegraphics[width=\textwidth]{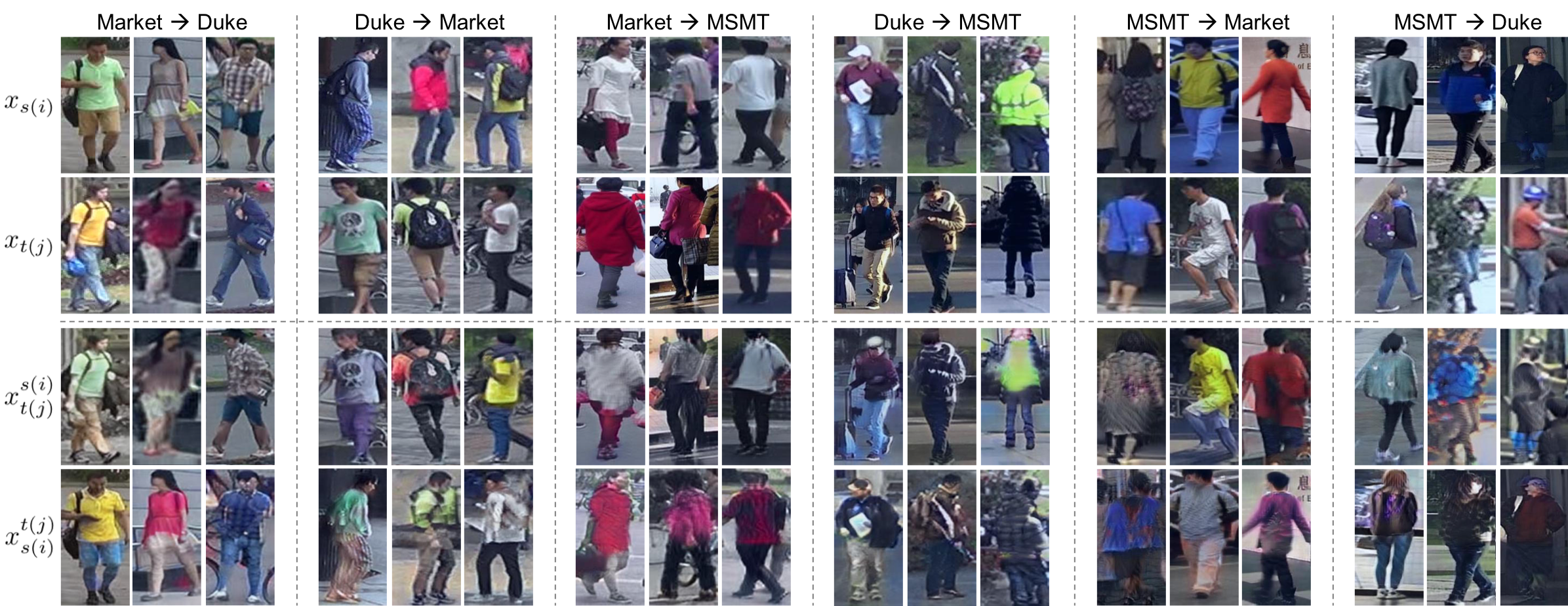}
	\caption{Examples of our synthesized images on six cross-domain benchmark pairs. We  show source images in the first row, target images in the second row, synthetic images with source appearance and target structure in the third row, and synthetic images with target appearance and source structure in the fourth row.}
    \vspace{-4pt}
	\label{fig:six_groups}
\end{figure*}

\noindent\textbf{Comparison with the state-of-the-art.} We also compare the image generation results between DG-Net++ and other representative image translation based methods for unsupervised cross-domain person re-id, including CycleGAN \cite{zhu2017unpaired}, SPGAN \cite{wei2018person}, PNA-Net \cite{li2019cross} and CSGLP \cite{ren2019domain}. As shown in Figure \ref{fig:translation}, CycleGAN and SPGAN virtually translate the illumination only. CSGLP can switch the illumination and background between two domains, but is not able to change foreground or person appearance. PDA-Net synthesizes various images by manipulating human poses, but the generated images are prone to be blurry. In comparison, our generated images look more realistic in terms of both foreground and background. This also verifies the effectiveness of the proposed framework to decompose id-related and id-unrelated factors, and therefore facilitating more effective cross-domain adaptation.

\noindent\textbf{Cross-domain synthesized images.} Here we show more qualitative results of cross-domain generated images in Figure~\ref{fig:six_groups}, which shows the examples on six cross-domain pairs. Compared to the within-domain image generation~\cite{ge2018fd,pose-normalized,zheng2019joint}, the cross-domain image synthesis is more challenging due to huge domain gap and lack of identity supervision in target domain. DG-Net++ is able to generate realistic images over different domain pairs, which present very diverse clothing styles, seasons, poses, viewpoints, backgrounds, illuminations, etc. This indicates that our approach is not just geared to solve a particular type of domain gap but is generalizable across different domains. The last column of this figure shows a failure case where the source and target appearances are not well retained in the synthetic images. We conjecture that this difficulty is caused by the occluded bottom right person in the target image as his appearance confuses the appearance feature extraction. 


\noindent\textbf{Ablation study.} We then qualitatively compare our full model DG-Net++ to its two variants without cross-domain disentangling and pseudo-identity supervision. As shown in Figure~\ref{fig:visual_ablation}, removing cross-domain disentangling or further pseudo-id, the synthetic images are unsatisfying as the models fail to translate the accurate clothing color or style. This again clearly shows the merits of our unified disentangling and adaptation for cross-domain image generation.

\section{Conclusion}

In this paper, we have proposed a joint learning framework that disentangles id-related/unrelated factors and performs adaptation exclusively on the id-related feature space. This design leads to more effective adaptation as the id-unrelated noises are segregated from the adaptation process. Our cross-domain cycle-consistent image generation as well as adversarial alignment and self-training are co-designed such that the disentangling and adaptation modules can mutually promote each other during joint training. Experimental results on the six benchmarks find that our approach consistently brings substantial performance gains. We hope the proposed approach would inspire more work of integrating disentangling and adaptation for unsupervised cross-domain person re-id.

\clearpage
%
%
\bibliographystyle{splncs04}
\bibliography{egbib}

\begin{thebibliography}{10}
\providecommand{\url}[1]{\texttt{#1}}
\providecommand{\urlprefix}{URL }
\providecommand{\doi}[1]{https://doi.org/#1}

\bibitem{chen2020angular}
Chen, B., Liu, W., Yu, Z., Kautz, J., Shrivastava, A., Garg, A., Anandkumar,
  A.: Angular visual hardness. In: ICML (2020)

\bibitem{chen2020automated}
Chen, W., Yu, Z., Wang, Z., Anandkumar, A.: Automated synthetic-to-real
  generalization. In: ICML (2020)

\bibitem{chen2016infogan}
Chen, X., Duan, Y., Houthooft, R., Schulman, J., Sutskever, I., Abbeel, P.:
  {InfoGAN}: Interpretable representation learning by information maximizing
  generative adversarial nets. In: NeurIPS (2016)

\bibitem{yanbei-chen}
Chen, Y., Zhu, X., Gong, S.: Instance-guided context rendering for cross-domain
  person re-identification. In: ICCV (2019)

\bibitem{imagenet}
Deng, J., Dong, W., Socher, R., Li, L.J., Li, K., Fei-Fei, L.: {ImageNet}: A
  large-scale hierarchical image database. In: CVPR (2009)

\bibitem{deng2018image}
Deng, W., Zheng, L., Ye, Q., Kang, G., Yang, Y., Jiao, J.: Image-image domain
  adaptation with preserved self-similarity and domain-dissimilarity for person
  re-identification. In: CVPR (2018)

\bibitem{dundar2020domain}
Dundar, A., Liu, M.Y., Yu, Z., Wang, T.C., Zedlewski, J., Kautz, J.: Domain
  stylization: A fast covariance matching framework towards domain adaptation.
  TPAMI  (2020)

\bibitem{eom2019learning}
Eom, C., Ham, B.: Learning disentangled representation for robust person
  re-identification. In: NeurIPS (2019)

\bibitem{ester1996density}
Ester, M., Kriegel, H.P., Sander, J., Xu, X.: A density-based algorithm for
  discovering clusters in large spatial databases with noise. In: KDD (1996)

\bibitem{Fan_2020_CVPR}
Fan, L., Li, T., Fang, R., Hristov, R., Yuan, Y., Katabi, D.: Learning longterm
  representations for person re-identification using radio signals. In: CVPR
  (2020)

\bibitem{Fu_2019_ICCV}
Fu, Y., Wei, Y., Wang, G., Zhou, Y., Shi, H., Huang, T.S.: Self-similarity
  grouping: A simple unsupervised cross domain adaptation approach for person
  re-identification. In: ICCV (2019)

\bibitem{Ge2020Mutual}
Ge, Y., Chen, D., Li, H.: Mutual mean-teaching: Pseudo label refinery for
  unsupervised domain adaptation on person re-identification. In: ICLR (2020)

\bibitem{ge2018fd}
Ge, Y., Li, Z., Zhao, H., Yin, G., Yi, S., Wang, X., et~al.: {FD-GAN}:
  Pose-guided feature distilling {GAN} for robust person re-identification. In:
  NeurIPS (2018)

\bibitem{hadad2018two}
Hadad, N., Wolf, L., Shahar, M.: A two-step disentanglement method. In: CVPR
  (2018)

\bibitem{he2016deep}
He, K., Zhang, X., Ren, S., Sun, J.: Deep residual learning for image
  recognition. In: CVPR (2016)

\bibitem{hermans2017defense}
Hermans, A., Beyer, L., Leibe, B.: In defense of the triplet loss for person
  re-identification. arXiv:1703.07737  (2017)

\bibitem{higgins2017beta}
Higgins, I., Matthey, L., Pal, A., Burgess, C., Glorot, X., Botvinick, M.,
  Mohamed, S., Lerchner, A.: beta-{VAE}: Learning basic visual concepts with a
  constrained variational framework. In: ICLR (2017)

\bibitem{hoffman2018cycada}
Hoffman, J., Tzeng, E., Park, T., Zhu, J.Y., Isola, P., Saenko, K., Efros,
  A.A., Darrell, T.: {CyCADA}: Cycle-consistent adversarial domain adaptation.
  In: ICML (2018)

\bibitem{hong2018conditional}
Hong, W., Wang, Z., Yang, M., Yuan, J.: Conditional generative adversarial
  network for structured domain adaptation. In: CVPR (2018)

\bibitem{huang2017arbitrary}
Huang, X., Belongie, S.: Arbitrary style transfer in real-time with adaptive
  instance normalization. In: ICCV (2017)

\bibitem{Huang_2018_ECCV}
Huang, X., Liu, M.Y., Belongie, S., Kautz, J.: Multimodal unsupervised
  image-to-image translation. In: ECCV (2018)

\bibitem{huang2018multi}
Huang, Y., Xu, J., Wu, Q., Zheng, Z., Zhang, Z., Zhang, J.: Multi-pseudo
  regularized label for generated data in person re-identification. TIP  (2018)

\bibitem{ioffe2015batch}
Ioffe, S., Szegedy, C.: Batch normalization: Accelerating deep network training
  by reducing internal covariate shift. In: ICML (2015)

\bibitem{isola2017image}
Isola, P., Zhu, J.Y., Zhou, T., Efros, A.: Image-to-image translation with
  conditional adversarial networks. In: CVPR (2017)

\bibitem{semantic-parsing}
Kalayeh, M., Basaran, E., Muhittin~Gokmen, M.K., Shah, M.: Human semantic
  parsing for person re-identification. In: CVPR (2018)

\bibitem{kingma2014adam}
Kingma, D., Ba, J.: Adam: A method for stochastic optimization. In: ICLR (2015)

\bibitem{lee2019sliced}
Lee, C.Y., Batra, T., Baig, M.H., Ulbricht, D.: Sliced wasserstein discrepancy
  for unsupervised domain adaptation. In: CVPR (2019)

\bibitem{lee2018diverse}
Lee, H.Y., Tseng, H.Y., Huang, J.B., Singh, M., Yang, M.H.: Diverse
  image-to-image translation via disentangled representations. In: ECCV (2018)

\bibitem{li2019cross}
Li, Y.J., Lin, C.S., Lin, Y.B., Wang, Y.C.: Cross-dataset person
  re-identification via unsupervised pose disentanglement and adaptation. In:
  ICCV (2019)

\bibitem{li2018adaptation}
Li, Y.J., Yang, F.E., Liu, Y.C., Yeh, Y.Y., Du, X., Wang, Y.C.: Adaptation and
  re-identification network: An unsupervised deep transfer learning approach to
  person re-identification. In: CVPR Workshop (2018)

\bibitem{linmulti}
Lin, S., Li, H., Li, C.T., Kot, A.C.: Multi-task mid-level feature alignment
  network for unsupervised cross-dataset person re-identification. In: BMVC
  (2018)

\bibitem{adaptive-transfer}
Liu, J., Zha, Z.J., Chen, D., Hong, R., Wang, M.: Adaptive transfer network for
  cross-domain person re-identification. In: CVPR (2019)

\bibitem{liu2018pose}
Liu, J., Ni, B., Yan, Y., Zhou, P., Cheng, S., Hu, J.: Pose transferrable
  person re-identification. In: CVPR (2018)

\bibitem{long2015learning}
Long, M., Cao, Y., Wang, J., Jordan, M.I.: Learning transferable features with
  deep adaptation networks. In: ICML (2015)

\bibitem{tsne}
van~der Maaten, L., Hinton, G.: Visualizing high-dimensional data using t-sne.
  JMLR  (2008)

\bibitem{mathieu2016disentangling}
Mathieu, M.F., Zhao, J.J., Zhao, J., Ramesh, A., Sprechmann, P., LeCun, Y.:
  Disentangling factors of variation in deep representation using adversarial
  training. In: NeurIPS (2016)

\bibitem{aicity20}
Naphade, M., Wang, S., Anastasiu, D., Tang, Z., Chang, M.C., Yang, X., Zheng,
  L., Sharma, A., Chellappa, R., Chakraborty, P.: The 4th {AI} city challenge.
  In: CVPR Workshop (2020)

\bibitem{Qi_2019_ICCV}
Qi, L., Wang, L., Huo, J., Zhou, L., Shi, Y., Gao, Y.: A novel unsupervised
  camera-aware domain adaptation framework for person re-identification. In:
  ICCV (2019)

\bibitem{pose-normalized}
Qian, X., Fu, Y., Xiang, T., Wang, W., Qiu, J., Wu, Y., Jiang, Y.G., Xue, X.:
  Pose-normalized image generation for person re-identification. In: ECCV
  (2018)

\bibitem{ren2019domain}
Ren, C.X., Liang, B.H., Lei, Z.: Domain adaptive person re-identification via
  camera style generation and label propagation. arXiv:1905.05382  (2019)

\bibitem{dukemtmc}
Ristani, E., Solera, F., Zou, R., Cucchiara, R., Tomasi, C.: Performance
  measures and a data set for multi-target, multi-camera tracking. In: ECCV
  Workshop (2016)

\bibitem{song2018unsupervised}
Song, L., Wang, C., Zhang, L., Du, B., Zhang, Q., Huang, C., Wang, X.:
  Unsupervised domain adaptive re-identification: Theory and practice.
  arXiv:1807.11334  (2018)

\bibitem{su2017pose}
Su, C., Li, J., Zhang, S., Xing, J., Gao, W., Tian, Q.: Pose-driven deep
  convolutional model for person re-identification. In: ICCV (2017)

\bibitem{su2016deep}
Su, C., Zhang, S., Xing, J., Gao, W., Tian, Q.: Deep attributes driven
  multi-camera person re-identification. In: ECCV (2016)

\bibitem{sun2016deep}
Sun, B., Saenko, K.: Deep coral: Correlation alignment for deep domain
  adaptation. In: ECCV (2016)

\bibitem{sun2017svdnet}
Sun, Y., Zheng, L., Deng, W., Wang, S.: {SVDNet} for pedestrian retrieval. In:
  ICCV (2017)

\bibitem{sun2018beyond}
Sun, Y., Zheng, L., Yang, Y., Tian, Q., Wang, S.: Beyond part models: Person
  retrieval with refined part pooling (and a strong convolutional baseline).
  In: ECCV (2018)

\bibitem{pamtri}
Tang, Z., Naphade, M., Birchfield, S., Tremblay, J., Hodge, W., Kumar, R.,
  Wang, S., Yang, X.: {PAMTRI}: Pose-aware multi-task learning for vehicle
  re-identification using randomized synthetic data. In: ICCV (2019)

\bibitem{Tang_2019_CVPR}
Tang, Z., Naphade, M., Liu, M.Y., Yang, X., Birchfield, S., Wang, S., Kumar,
  R., Anastasiu, D., Hwang, J.N.: {CityFlow}: A city-scale benchmark for
  multi-target multi-camera vehicle tracking and re-identification. In: CVPR
  (2019)

\bibitem{tzeng2017adversarial}
Tzeng, E., Hoffman, J., Saenko, K., Darrell, T.: Adversarial discriminative
  domain adaptation. In: CVPR (2017)

\bibitem{wang2018transferable}
Wang, J., Zhu, X., Gong, S., Li, W.: Transferable joint attribute-identity deep
  learning for unsupervised person re-identification. In: CVPR (2018)

\bibitem{wei2018person}
Wei, L., Zhang, S., Gao, W., Tian, Q.: Person transfer {GAN} to bridge domain
  gap for person re-identification. In: CVPR (2018)

\bibitem{wei2017glad}
Wei, L., Zhang, S., Yao, H., Gao, W., Tian, Q.: Glad: Global-local-alignment
  descriptor for pedestrian retrieval. In: ACM Multimedia (2017)

\bibitem{xu2015empirical}
Xu, B., Wang, N., Chen, T., Li, M.: Empirical evaluation of rectified
  activations in convolutional network. In: ICML Workshop (2015)

\bibitem{yang2019patch}
Yang, Q., Yu, H.X., Wu, A., Zheng, W.S.: Patch-based discriminative feature
  learning for unsupervised person re-identification. In: CVPR (2019)

\bibitem{vehiclex}
Yao, Y., Zheng, L., Yang, X., Naphade, M., Gedeon, T.: Simulating content
  consistent vehicle datasets with attribute descent. In: ECCV (2020)

\bibitem{zheng2015scalable}
Zheng, L., Shen, L., Tian, L., Wang, S., Wang, J., Tian, Q.: Scalable person
  re-identification: A benchmark. In: ICCV (2015)

\bibitem{attentive-siamese}
Zheng, M., Karanam, S., Wu, Z., Radke, R.: Re-identification with consistent
  attentive siamese networks. In: CVPR (2019)

\bibitem{zheng2019joint}
Zheng, Z., Yang, X., Yu, Z., Zheng, L., Yang, Y., Kautz, J.: Joint
  discriminative and generative learning for person re-identification. In: CVPR
  (2019)

\bibitem{zheng2020person}
Zheng, Z., Yang, Y.: Person re-identification in the {3D} space. arXiv
  2006.04569  (2020)

\bibitem{unlabeld-samples}
Zheng, Z., Zheng, L., Yang, Y.: Unlabeled samples generated by {GAN} improve
  the person re-identification baseline in vitro. In: ICCV (2017)

\bibitem{zheng2018pedestrian}
Zheng, Z., Zheng, L., Yang, Y.: Pedestrian alignment network for large-scale
  person re-identification. TCSVT  (2018)

\bibitem{zhong2017re}
Zhong, Z., Zheng, L., Cao, D., Li, S.: Re-ranking person re-identification with
  k-reciprocal encoding. In: CVPR (2017)

\bibitem{zhunzhong2018eccv}
Zhong, Z., Zheng, L., Li, S., Yang, Y.: Generalizing a person retrieval model
  hetero- and homogeneously. In: ECCV (2018)

\bibitem{zhong2019invariance}
Zhong, Z., Zheng, L., Luo, Z., Li, S., Yang, Y.: Invariance matters: Exemplar
  memory for domain adaptive person re-identification. In: CVPR (2019)

\bibitem{zhu2017unpaired}
Zhu, J.Y., Park, T., Isola, P., Efros, A.A.: Unpaired image-to-image
  translation using cycle-consistent adversarial networks. In: ICCV (2017)

\bibitem{zou2018unsupervised}
Zou, Y., Yu, Z., Kumar, B.V., Wang, J.: Unsupervised domain adaptation for
  semantic segmentation via class-balanced self-training. In: ECCV (2018)

\bibitem{Zou_2019_ICCV}
Zou, Y., Yu, Z., Liu, X., Kumar, B.V., Wang, J.: Confidence regularized
  self-training. In: ICCV (2019)

\end{thebibliography}

\clearpage

\section*{Appendix}

\setcounter{table}{2}
\setcounter{figure}{6}

\appendix

\section{Additional Implementation Details}

DG-Net++ consists of an appearance encoder $E_{app}$, two source and target structure encoders $E_{str}^{s}, E_{str}^{t}$, two source and target decoders $G^s, G^t$, an image discriminator $D_{img}$, and a domain discriminator $D_{dom}$. As described in the main paper, $E_{app}$ is modified from ResNet50, and $E_{str}^{s}, E_{str}^{t}$ and $G^s, G^t$ follow the within-domain architecture designs as DG-Net \cite{zheng2019joint}. $D_{dom}$ is a multi-layer perceptron containing four fully-connected layer, where the input dimension is $2048$, output dimension is 1, and the dimensions of hidden layers are $1024$, $512$ and $256$. Note after each fully connected layer, we apply a batch normalization layer \cite{ioffe2015batch} and a LReLU \cite{xu2015empirical} (negative slope set to 0.2). In all experiments, the input images are resized to $256\times128$. SGD is used to train $E_{app}$ with learning rate $0.0006$ and momentum $0.9$, and Adam is applied to optimize $E_{str}^s$, $E_{str}^t$, $G^s$, $G^t$, $D_{img}$ with learning rate $0.000001$ and $(\beta_1, \beta_2) = (0, 0.999)$, and $D_{dom}$ with learning rate $0.00001$. To warm up $E_{app}$, $E_{str}^s$, $G^s$ and $D_{img}$, we follow the configuration as~\cite{zheng2019joint}. 
We use an iterative self-training approach to generate pseudo-labels every two epochs. We utilize labeled source and pseudo-labeled target data in self-training with softmax loss. DBSCAN is used for clustering with k-reciprocal encoding to compute pairwise distances. 
Every experiment is conducted on a single NVIDIA TITAN V100 GPU. Our full model takes 15.8 GPU memory and runs for 460K iterations. Our source code with all implementation details is available at~\url{https://github.com/NVlabs/DG-Net-PP}. 

\section{Feature Distribution Visualization}

\begin{figure}[t!]
\centering     
\subfigure[]{\label{fig:hyper}\includegraphics[width=0.49\textwidth]{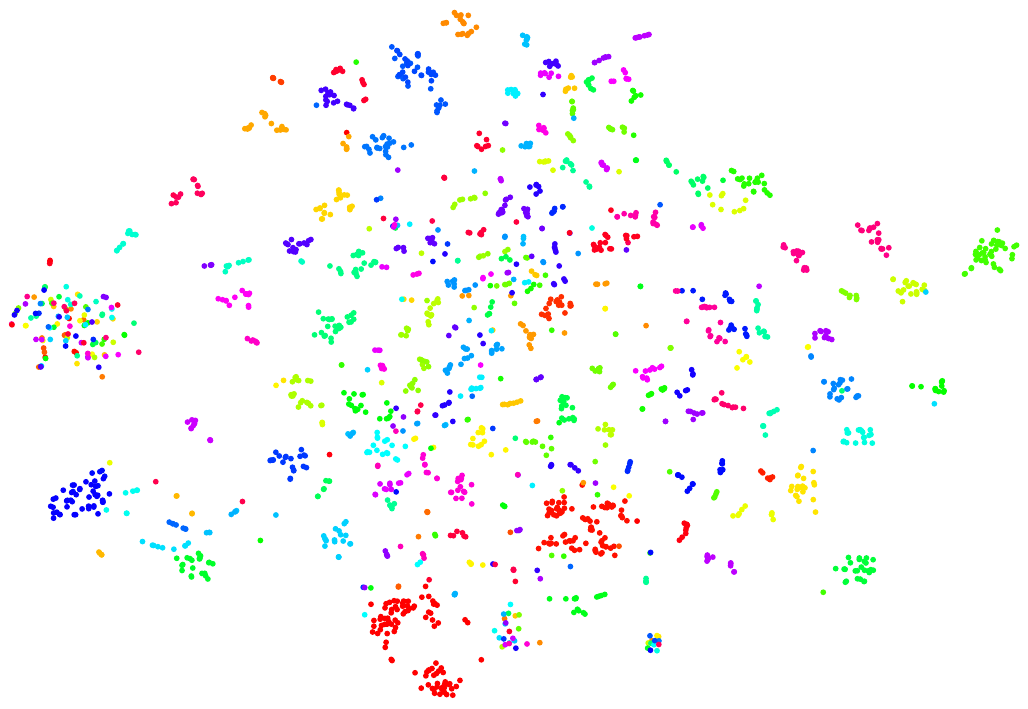}}~~
\subfigure[]{\label{fig:visual_ablation}\includegraphics[width=0.49\textwidth]{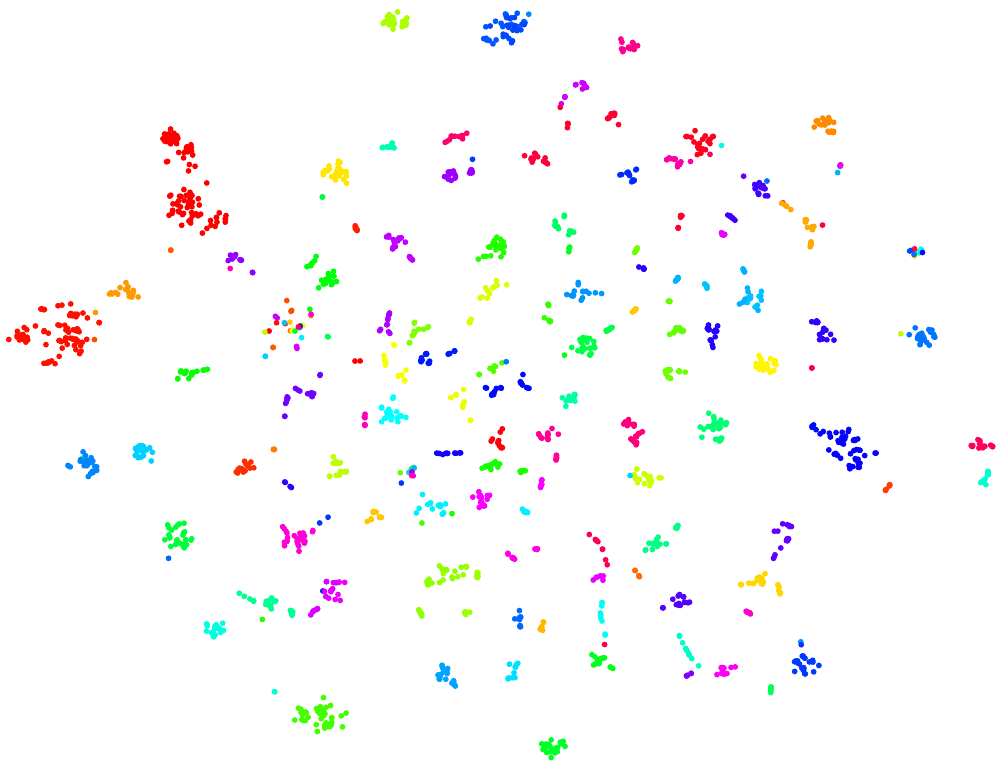}}
\vspace{-.2cm}
\caption{Visualization by t-SNE of the feature distributions of the images in target domain. Features are extracted by (a) the baseline method and (b) DG-Net++ on the cross-domain pair Market $\rightarrow$ Duke.}
\label{fig:tsne}
\vspace{-5pt}
\end{figure}

DG-Net++ is a joint learning framework that disentangles id-related/unrelated factors such that adaptation can be more effectively conducted on id-related space to prevent id-unrelated interference. Figure~\ref{fig:tsne} illustrates the feature distributions of the images in target domain visualized by t-SNE~\cite{tsne}. It can be apparently observed that by using DG-Net++ the features of different identities are more separable and the features of the same identity are more clustered.          
To further quantitatively evaluate the target domain feature distributions, we compute the purity scores for the features produced by the baseline method and DG-Net++ on the cross-domain pair Market $\rightarrow$ Duke. 
To compute the purity score, each cluster is assigned to the identity that is most frequent in the cluster, then the purity score is measured by the number of correctly assigned images divided by the total number of images. The purity score is 51.9\% for baseline and 76.3\% for DG-Net++, clearly indicating that the intra-class similarity and inter-class difference are more encouraged in DG-Net++.

\section{Additional Ablation Study}
\begin{table}[t]
\centering
\begin{tabular}{cccccccccc}
\shline
\multicolumn{10}{c}{Market-1501 $\rightarrow$ DukeMTMC-reID} \\ \hline
\multicolumn{1}{c|}{$\varepsilon$} & 0.35 & 0.40 & 0.45 & \multicolumn{1}{c||}{0.50} & \multicolumn{1}{c|}{\texttt{MinPts}} & 5 & 6 & 7 & 8 \\
\multicolumn{1}{c|}{mAP} & 62.4 & 61,7 & 63.8 & \multicolumn{1}{c||}{62.2} & \multicolumn{1}{c|}{mAP} & 61.4 & 61.6 & 63.8 & 62.8 \\ \shline
\end{tabular}%
\vspace{1mm}
\caption{Sensitivity analysis of the hyper-parameters $\varepsilon$ (the maximum distance between two samples to be treated as neighbors) and \texttt{MinPts} (the minimal number of neighboring samples of a point to be selected as a core point) of DBSCAN.}
\label{DBSCAN-SA}
\vspace{-24pt}
\end{table}

\begin{table}[b]
\vspace{-12pt}
\centering
\resizebox{\linewidth}{!}{%
\begin{tabular}{l|c|c|c|c|c|c|c}
\shline
\multirow{2}{*}{Method} & \multirow{2}{*}{Metric} & Market$\rightarrow$ & Duke$\rightarrow$ & MSMT$\rightarrow$ & Market$\rightarrow$ & MSMT$\rightarrow$ & Duke$\rightarrow$ \\
 &  & Duke & Market & Market & MSMT & Duke & MSMT \\ \hline
\multirow{2}{*}{DG-Net~\cite{zheng2019joint}} & Rank@1 & 42.6 & 56.1 & 61.8 & 17.1 & 61.9 & 20.6 \\
 & mAP & 24.3 & 26.8 & 33.6 & 5.4 & 40.7 & 6.4 \\ \hline
\multirow{2}{*}{DG-Net++} & Rank@1 & \textbf{78.9} (+36.3) & \textbf{82.1} (+26.0) & \textbf{83.1} (+21.3) & \textbf{48.4} (+31.3) & \textbf{75.2} (+13.3) & \textbf{48.8} (+28.2) \\
 & mAP & \textbf{63.8} (+39.5) & \textbf{61.7} (+34.9) & \textbf{64.6} (+31.0) & \textbf{22.1} (+16.7) & \textbf{58.2} (+17.5) & \textbf{22.1} (+15.7) \\ \shline
\end{tabular}%
}
\vspace{1mm}
\caption{Comparison between DG-Net and DG-Net++ for unsupervised cross-domain person re-id on the six benchmark pairs.}
\label{DG-DG++}
\vspace{-12pt}
\end{table}

\noindent{\textbf{Sensitivity analysis of DBSCAN.}} We adopt DBSCAN to produce the pseudo-labels of images in target domain. Experiments show that our model is not sensitive to the hyper-parameters of DBSCAN. Specifically, we conduct sensitivity analysis for (1) $\varepsilon$ which is the maximum distance between two samples to be considered as neighbors, and (2) \texttt{MinPts} which is the minimal number of neighbouring samples for a point to be considered as a core point. Table~\ref{DBSCAN-SA} shows the experimental results for sensitivity analysis of $\varepsilon$ (fixing \texttt{MinPts} to 7) and \texttt{MinPts} (fixing $\varepsilon$ to $0.45$) on the benchmark pair Market $\rightarrow$ Duke. It can be found that DG-Net++ is overall not sensitive to $\varepsilon$ and \texttt{MinPts}.

\noindent{\textbf{DG-Net++ vs. DG-Net.}} To illustrate the cross-domain performance difference between DG-Net~\cite{zheng2019joint} and DG-Net++, we show their comparisons over the six cross-domain pairs in Table~\ref{DG-DG++}. DG-Net++ is found to substantially and consistently outperform DG-Net over all benchmarks. This is evident to validate the efficacy of the proposed learning framework in coupling cross-domain disentanglement and adaptation, backing the necessity of such combination for unsupervised cross-domain re-id. 

\end{document}